\documentclass{article}
\PassOptionsToPackage{numbers,sort&compress}{natbib}
\usepackage[T1]{fontenc} 
\usepackage{iclr2027_conference,times}
\setcitestyle{numbers,square,comma,sort&compress}

\usepackage{amsmath,amsfonts,bm}

\def\eqref#1{equation~\ref{#1}}

\def\1{\bm{1}}

\DeclareMathAlphabet{\mathsfit}{\encodingdefault}{\sfdefault}{m}{sl}
\SetMathAlphabet{\mathsfit}{bold}{\encodingdefault}{\sfdefault}{bx}{n}

\usepackage{pifont}
\usepackage{graphicx}
\usepackage{booktabs}
\usepackage{multirow}
\usepackage{tabularx}
\usepackage[section]{placeins} 
\newcolumntype{Y}{>{\centering\arraybackslash}X}

\usepackage{hyperref}
\usepackage{url}
\usepackage{xcolor}
\usepackage{colortbl}
\usepackage[skins]{tcolorbox}
\definecolor{EgoProbeCover}{HTML}{E8F1FA}
\newtcolorbox{arxivcover}{
  enhanced, colback=EgoProbeCover, colframe=EgoProbeCover,
  boxrule=0pt, arc=9pt,
  left=12pt, right=12pt, top=12pt, bottom=12pt,
  before skip=0pt, after skip=0pt
}
\makeatletter
\newcommand{\arxivtitleblock}{%
  {\sffamily\bfseries\fontsize{15}{18}\selectfont\raggedright\@title\par}
  \vspace{9pt}
  {\sffamily\small\begin{tabular}{@{}l@{}}\@author\end{tabular}\par}
}
\makeatother
\renewenvironment{abstract}{%
  \par\vspace{12pt}\normalfont\normalsize
  \setlength{\parindent}{0pt}\setlength{\parskip}{0pt}
}{\par}

\newcommand{\fyq}[1]{\textcolor{black}{#1}}
\newcommand{\todo}[1]{\textcolor{black}{#1}}

\title{EgoProbe: Benchmarking Long-Range and Exact Memory in Streaming Egocentric Videos}
\author{%
\textbf{Haokai Pang}$^{1*}$ \enspace
\textbf{Yuqian Fu}$^{2*\dagger}$ \enspace
\textbf{Yifei Wang}$^{3}$ \enspace
\textbf{Yujiao Shen}$^{4}$ \enspace
\textbf{Tianwen Qian}$^{3}$ \\
\textbf{Mohamed Elhoseiny}$^{2}$ \enspace
\textbf{Hongbo Fu}$^{1}$ \\[0.5em]
{\normalfont\small $^{1}$HKUST} \enspace
{\normalfont\small $^{2}$KAUST} \enspace
{\normalfont\small $^{3}$ECNU \quad $^{4}$NTU}
}
\iclrfinalcopy

\begin{document}
\raggedbottom 
\fancyhead{}
\renewcommand{\headrulewidth}{0pt}
\begin{arxivcover}
\arxivtitleblock

\begin{abstract}

\fyq{Egocentric video provides a continuous first-person record of everyday activities, making it a natural foundation for wearable AI assistants. Unlike offline video understanding, streaming systems must process observations as they arrive, while queries may refer to experiences from the distant past. 
Recent work has begun to explore whether information can be recalled across long temporal gaps. Yet remembering how far is only part of the challenge: recurring objects, actions, and locations also require models to remember exactly which detail belongs to which event. 
We introduce \textbf{EgoProbe}, a benchmark that evaluates streaming egocentric memory along these two complementary dimensions: long-range retention, i.e., \textit{how far}, and event-specific recall under interference, i.e., \textit{how exact}. 
Built on long-form egocentric videos, EgoProbe uses an event-grounded data engine to extend recall intervals, identify related events, and construct event-grounded distractors, followed by human verification. It contains two subsets: \textbf{EgoProbe-Base}, which evaluates long-range recall without intervening related events, and \textbf{EgoProbe-Echo}, which further challenges models with similar intervening events and event-grounded distractors. EgoProbe-Echo additionally adopts coarse-to-fine multi-turn probing: the first turn evaluates coarse-grained event recall, while the second tests whether the model can retrieve the target-specific detail associated with the designated event.
\todo{Extensive evaluation across diverse models reveals substantial limitations in long-range recall, with further challenges in recovering target-event details among related experiences. Even when models correctly answer coarse-grained questions, they often fail to recall the specific details of the designated event. EgoProbe exposes this underexplored failure mode and provides a testbed for advancing long-range and exact memory in streaming egocentric video understanding. Code and data will be released.}
}

\end{abstract}
\end{arxivcover}
\begingroup
\renewcommand{\thefootnote}{}
\footnotetext{$^{*}$Equal contribution. $^{\dagger}$Corresponding author.}
\endgroup

\par\vspace{16pt}
\let\arxivoriginalsection\section
\renewcommand{\section}[1]{%
  \arxivoriginalsection{#1}%
  \let\section\arxivoriginalsection
  \suppressfloats[t]%
\begin{figure}[t]
\centering
\includegraphics[width=1\linewidth]{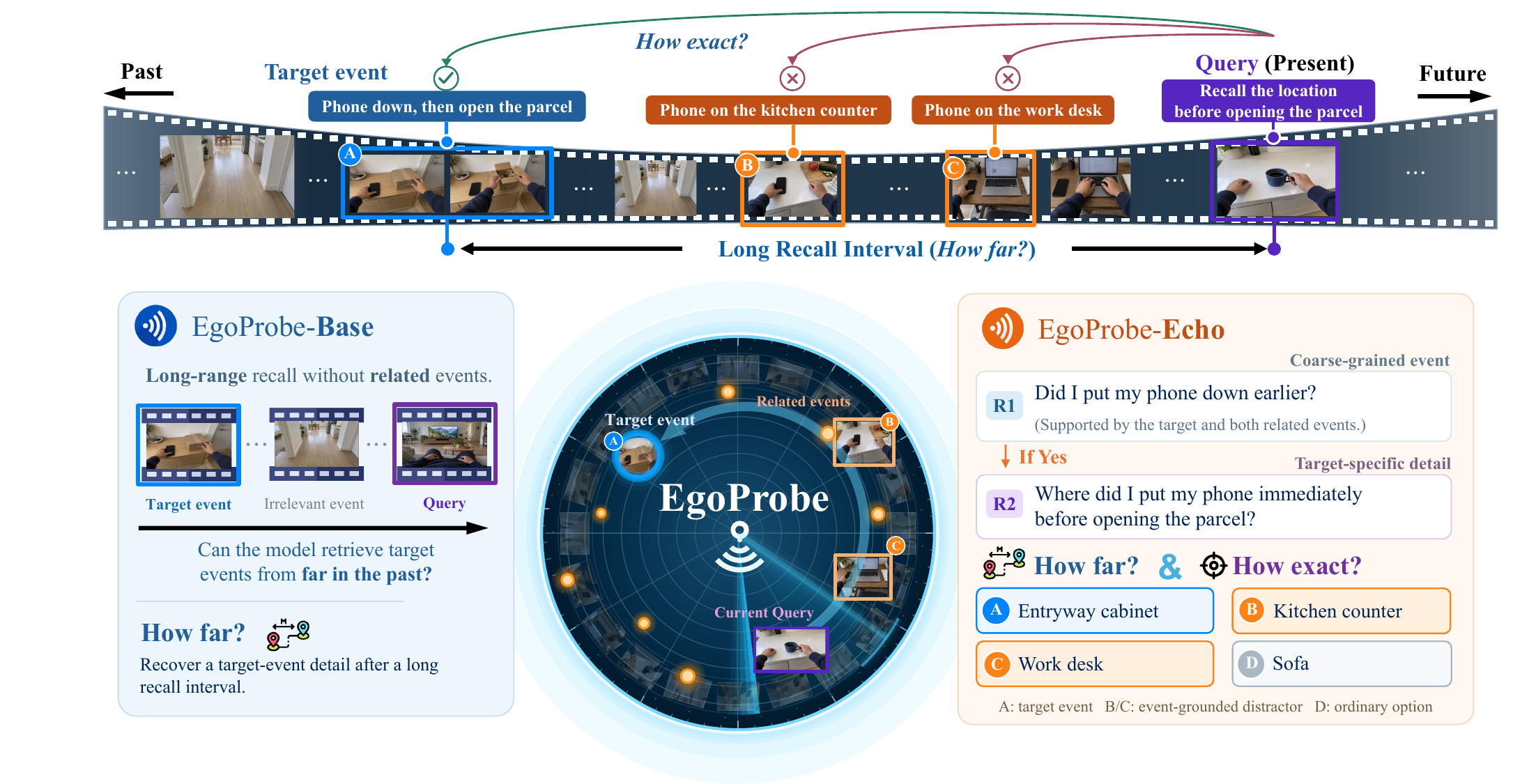}
\caption{\textbf{Overview of EgoProbe.} EgoProbe-Base probes long-range recall without intervening related events. EgoProbe-Echo further uses coarse-to-fine questions and event-grounded distractors to probe recall of target-event details among related experiences.}
\label{fig:teaser}
\end{figure}
}
\section{Introduction}
\label{sec:introduction}
\fyq{
Wearable AI assistants, e.g., always-on intelligent glasses, are expected to perceive continuously as users observe and interact with the world ~\cite{grauman2022ego4d, damen2020epic}. In such real-world settings, useful information is scattered across a long, evolving visual stream rather than confined to a short and self-contained video clip. A query raised at the current moment may depend on an object seen, an action performed, or an interaction that occurred minutes or even hours earlier~\cite{barmann2022did}. Supporting such assistance therefore requires models to move beyond local perception to long-range memory for egocentric video streams.
}

\fyq{
Despite growing interest in streaming video understanding, existing benchmarks ~\cite{niu2025ovo, lin2026streamingbench, fu2025video} remain limited in three important aspects. First, many benchmarks~\cite{li2024mvbench, fu2025video} are designed for generic video rather than the specific demands of egocentric streams, which present distinctive challenges including restricted fields of view, substantial ego-motion, and frequent interactions with surrounding.
Second, a streaming protocol alone does not necessarily imply long-range memory dependence: many queries may still be answered from recent observations, without requiring information to be retained across long recall intervals~\cite{shen2026simple}. Third, recent work~\cite{forte2026egostream} has begun to study long-range memory in egocentric streaming settings, largely focusing on how far information can be recalled. However, temporal distance alone does not fully characterize reliable memory. In everyday egocentric streams, similar events may recur, requiring models not only to retain information over time but also to associate each recalled detail with the correct event.
}

\fyq{
These limitations naturally motivate the design of \textbf{EgoProbe} (Fig.~\ref{fig:teaser}), which is designed to 1) reflect the distinctive characteristics of \textit{egocentric streams}; 2) explicitly require \textit{long-range} memory; and 3) also enable the evaluation of \textit{how exact} the models can recall details of the target event.
}

\fyq{
To achieve these goals, we build EgoProbe on two representative long-form egocentric video sources, Ego4D~\cite{grauman2022ego4d} and EgoLife~\cite{yang2025egolife}, which naturally provide the first-person, continuous visual streams required for egocentric evaluation and form a strong basis for studying long-range memory. Their associated episodic QA annotations, however, were not specifically designed to assess either long-range memory or precise recall of target-event details in the presence of related events. We therefore develop an event-grounded data engine to transform these questions into controlled probes of long-range memory and event-specific recall.
For \textit{long-range} memory, we use delayed query to control the recall interval between the target-event evidence and the query. Since additional related events before the delayed query may make the original question ambiguous, the engine further checks and, when necessary, disambiguates the question's reference to the target event to ensure that each extended question remains uniquely and correctly grounded in the video.
For \textit{exact} memory, we exploit recurring event patterns in egocentric streams to construct challenging distractors from related events that share similar semantics but differ in event-specific details. Rather than using arbitrary incorrect options, these distractors are grounded in actual observations from the same stream, allowing us to test whether a model recalls the correct detail of the target event without confusing it with details from related events. We further extend this evaluation into coarse-to-fine two-turn dialogues: the first turn asks about shared coarse event content, while the second queries the target-specific detail. This design enables EgoProbe to distinguish coarse-grained event recall from target-specific detail recall.
}

\fyq{
EgoProbe consists of two complementary subsets, both spanning various duration of recall intervals. \textbf{EgoProbe-Base} contains 1,137 single-turn questions without intervening related events; \textbf{EgoProbe-Echo} contains 1,229 single-turn questions with contrastive related events, together with 1,229 derived coarse-to-fine two-turn dialogues, enabling evaluation of event-specific memory under interference. All final QA pairs and their associated event evidence undergo human verification and refinement. Together, the two subsets support complementary analysis of long-range retention and event-specific recall in streaming egocentric video.
}

We evaluate a diverse range of advanced MLLMs across different model families. Our results reveal substantial gaps in both long-range memory and event-specific detail recall. On EgoProbe-Base, models struggle to recall distant target events even without intervening related experiences, with particularly low accuracy over ultra-long intervals. On EgoProbe-Echo, models frequently select details belonging to other related events, while successful coarse-grained responses do not ensure correct recall of target-specific details. These findings highlight the remaining challenges in retaining information over extended streams and recalling the correct details of a designated experience.

Our contributions are threefold:
(1) We introduce \textbf{EgoProbe}, a benchmark for long-range and event-specific memory in streaming egocentric video, combining recall-interval evaluation with event-grounded distractors and coarse-to-fine two-turn probing.
(2) We develop an event-grounded data engine and curate 2,366 independent QA pairs with 1,229 derived two-turn dialogues, supported by human verification of QA quality and associated event evidence.
(3) We extensively evaluate diverse model families, revealing limitations in long-range recall, errors associated with other related experiences, and gaps between coarse-grained responses and target-specific detail recall.

\section{Related Work}
\label{sec:related_work}

\subsection{Egocentric Video Understanding}
\label{sec:related_egocentric}

\fyq{
Egocentric video understanding has attracted growing attention due to its increasing importance for wearable AI assistants~\cite{zhou2024glassmail,zhang2023egospot,lee2024gazepointar} and embodied intelligence~\cite{wang2026oflow, lin2026evo,lin2026la4vla,wang2026afford}, giving rise to numerous datasets and benchmarks~\cite{grauman2022ego4d, damen2020epic, perrett2025hd, grauman2024ego, mangalam2023egoschema, xu2025tog, zhu2026egosound, zhang2026egonight}.
Representative datasets such as EPIC-KITCHENS~\cite{damen2020epic}, Ego4D~\cite{grauman2022ego4d}, HD-EPIC~\cite{perrett2025hd}, and Ego-Exo4D~\cite{grauman2024ego} support the study of everyday activities~\cite{song2023ego4d, wang2023ego}, human-object interactions~\cite{xu2023egopca, li2026egocentric}, and ego-exo relations~\cite{fu2025objectrelator, pan2026v, mahdi2025exo2egosyn}, while benchmarks such as EgoSchema~\cite{mangalam2023egoschema}, EgoTempo~\cite{plizzari2025omnia}, and EgoCross~\cite{li2026egocross} explore egocentric video understanding from different perspectives.
Beyond these tasks, growing attention has been devoted to understanding and recalling past experiences over extended periods.
QaEgo4D~\cite{barmann2022did}, derived from Ego4D ~\cite{grauman2022ego4d}, investigates episodic-memory question answering, while EgoLife~\cite{yang2025egolife} explores long-context understanding of prolonged egocentric recordings for everyday life assistance.
While these datasets provide valuable foundations for studying memory over extended first-person observations, their use in streaming settings remains comparatively underexplored, despite the growing need for always-on wearable assistants that continuously process observations and maintain access to past information. Building on Ego4D~\cite{grauman2022ego4d} and EgoLife~\cite{yang2025egolife}, EgoProbe investigates long-range and event-specific memory in this streaming setting.
}

\subsection{\fyq{Offline and Long Video Understanding}}
\label{sec:related_offline}

\fyq{With the rapid development of multimodal large language models (MLLMs), video understanding has advanced significantly.
Models such as LLaVA~\cite{li2024llava}, InternVL~\cite{wang2025internvl3}, and Qwen~\cite{team2025qwen3}, alongside proprietary models including Gemini~\cite{team2023gemini} and GPT~\cite{achiam2023gpt}, have demonstrated strong capabilities across diverse video understanding tasks.
Building upon these general-purpose MLLMs, recent approaches have further explored long-context video modeling.
For example, LongVA~\cite{zhang2024long} transfers long-context capabilities from language to vision, LongVU~\cite{shen2024longvu} employs adaptive spatiotemporal compression, and VideoChat-Flash~\cite{li2026videochat} introduces hierarchical compression for efficient long-video understanding.
Despite these advances, offline video understanding methods typically assume that the entire video is available at query time, whereas streaming systems must process observations as they arrive and maintain memory for future queries.}

\subsection{\fyq{Streaming Video Understanding and Benchmarks}}
\label{sec:related_streaming}
\fyq{
Recent streaming video approaches have explored several different perspectives. For example, VideoLLM-online ~\cite{chen2024videollm} develop streaming video dialogue systems, while Flash-VStream ~\cite{zhang2025flash} investigate memory mechanisms for continuous observation. StreamingVLM ~\cite{xu2026streamingvlm} and HERMES ~\cite{zhang2026hermes} explore KV-cache management and hierarchical memory, respectively. 
MEMO~\cite{li2026memo} and
ObjectStream~\cite{dong2026objectstream} further emphasize object-centric
memory for preserving fine-grained entity information across time.
StreamForest ~\cite{zeng2026streamforest}, OASIS ~\cite{liang2026oasis}, and EventMemAgent ~\cite{wen2026eventmemagent} focus on structured event memory and retrieval, while Video Streaming Thinking ~\cite{guan2026video} and JoyAI-VL-Interaction ~\cite{yao2026joyai} interleave reasoning with incoming observations.}

\fyq{Existing benchmarks, including StreamingBench ~\cite{lin2026streamingbench}, OVBench ~\cite{huang2025online}, OVO-Bench ~\cite{niu2025ovo}, and RIVER ~\cite{shi2026river}, evaluate online perception, contextual understanding, retrospective reasoning, and anticipation. More recent benchmarks explicitly investigate streaming memory and interaction. EGOSTREAM ~\cite{forte2026egostream} evaluates episodic memory across recall intervals and memory dimensions through answer-validity windows. EgoSAT ~\cite{lei2026egosat} covers retrospective, ongoing, and prospective interaction understanding, while S-EMBER ~\cite{wang2026s} evaluates grounded episodic retrieval with supporting temporal evidence.
Complementing these benchmarks, EgoProbe jointly evaluates long-range retention and exact event-specific recall using distractors grounded in related events. Its coarse-to-fine dialogues further distinguish coarse event recall from target-specific detail retrieval.
Table~\ref{tab:benchmark_comparison} summarizes the key differences between EgoProbe and related benchmarks.
}

\begin{table}[tbp]
\centering
\setlength{\abovecaptionskip}{0pt}
\caption{Comparison with related benchmarks. $\triangle$ denotes partial egocentric coverage; --- denotes unavailable comparable statistics. Event-grounded distractors link options to evidence from other related events. Video span includes recording gaps. $^{*}$EgoStream's span refers to its EgoLife source. $^{\dagger}$EgoLife reports recall intervals exceeding 24 h without specifying the exact maximum.}
\label{tab:benchmark_comparison}
\footnotesize
\setlength{\tabcolsep}{2pt}
\renewcommand{\arraystretch}{1.2}
\begin{tabularx}{\linewidth}{@{}>{\raggedright\arraybackslash}p{0.245\linewidth}
>{\hsize=0.99\hsize\linewidth=\hsize}Y
>{\hsize=0.99\hsize\linewidth=\hsize}Y
>{\hsize=0.84\hsize\linewidth=\hsize}Y
>{\hsize=1.34\hsize\linewidth=\hsize}Y
>{\hsize=0.97\hsize\linewidth=\hsize}Y
>{\hsize=0.90\hsize\linewidth=\hsize}Y
>{\hsize=0.97\hsize\linewidth=\hsize}Y@{}}
\toprule
\textbf{\scriptsize Benchmark} & \textbf{\scriptsize Egocentric} & \textbf{\scriptsize Streaming} & \textbf{\scriptsize\shortstack{Memory\\Recall}} & \textbf{\scriptsize\shortstack{Event-grounded\\Distractors}} & \textbf{\scriptsize\shortstack{Multi-turn\\Dialogue}} & \textbf{\scriptsize\shortstack{Max. Video\\Span}} & \textbf{\scriptsize\shortstack{Max. Recall\\Duration}} \\
\midrule
EgoTempo~\cite{plizzari2025omnia} & \ding{51} & \ding{55} & \ding{55} & \ding{55} & \ding{55} & 140 s & --- \\
EgoLife / EgoLifeQA~\cite{yang2025egolife} & \ding{51} & \ding{55} & \ding{51} & \ding{55} & \ding{55} & 7 days & $>24$ h$^{\dagger}$ \\
OVO-Bench~\cite{niu2025ovo} & $\triangle$ & \ding{51} & \ding{51} & \ding{55} & \ding{55} & 2.51 h & --- \\
RIVER~\cite{shi2026river} & $\triangle$ & \ding{51} & \ding{51} & \ding{55} & \ding{55} & 4.11 h & 1 h \\
StreamingBench~\cite{lin2026streamingbench} & \ding{55} & \ding{51} & \ding{51} & \ding{55} & \ding{51} & 24 min & --- \\
EgoStream~\cite{forte2026egostream} & \ding{51} & \ding{51} & \ding{51} & \ding{55} & \ding{55} & 7 days$^{*}$ & 45.3 h \\
\midrule
\textbf{EgoProbe (Ours)} & \ding{51} & \ding{51} & \ding{51} & \ding{51} & \ding{51} & \textbf{7 days} & \textbf{138.54 h} \\
\bottomrule
\end{tabularx}
\end{table}

\section{EgoProbe Benchmark}
\label{sec:benchmark}

\subsection{\fyq{Task Formulation and Core Design Principles}}
\label{sec:task_formulation}

\fyq{EgoProbe evaluates long-range memory and event-specific recall through retrospective question answering in streaming egocentric video. 
Given a streaming egocentric video $V$, a model processes incoming observations and continuously updates its memory without knowing future queries. At time $t_q$, the model receives a question $q$ about a past target event $e^\star$ and must retrieve the requested detail from its accumulated memory.}

\fyq{\textbf{Enabling How Far: Long-range memory evaluation.} To explicitly evaluate long-range memory, we define a controllable
\textit{recall interval}, $\Delta t = t_q - t_e^\star$, where
$t_e^\star$ denotes the end of the target-event evidence.
By adjusting the query arrival time, eligible questions can be
extended from short to substantially longer recall intervals,
requiring models to retrieve information observed much earlier
in the stream.} 
\fyq{
However, extending the recall interval may introduce additional events that make the original question ambiguous with respect to its target event.
We therefore apply \textit{query disambiguation} to ensure that each extended question uniquely identifies its target event while preserving the original answer.
}

\fyq{\textbf{Enabling How Exact: Event-specific recall evaluation.} To evaluate whether models can associate recalled details with the target event, we define \textit{related events} as distinct occurrences that share
a question-relevant semantic pattern with the target event, such
as a common action, object, or activity. We use alternative answers supported by these related events as \emph{event-grounded distractors}: details that are valid for another experience but incorrect for the target event. These distractors require models to recall the requested target-event detail without confusing it with details from similar experiences in the stream.} 
\fyq{As illustrated in Fig.~\ref{fig:teaser}, a user puts down a phone in an entryway cabinet before opening a parcel, and later places it on a kitchen counter and a work desk.
For the query ``Where did I put my phone before opening the parcel?'', only the first event is the target event, while the later locations serve as event-grounded distractors.
Correct recall therefore requires associating ``entryway cabinet'' with the designated event rather than recalling another valid phone location from the stream.
To further distinguish different levels of memory, we introduce \textit{coarse-to-fine two-turn probing}. The first round (R1) asks a yes/no question about the shared event pattern, e.g., ``Did I put my phone down?'' An affirmative response triggers the second round (R2), which asks for the detail associated with the designated target event, e.g., ``Where did I put it before opening the parcel?''}

\begin{figure}[tbp]
\centering
\includegraphics[width=\linewidth]{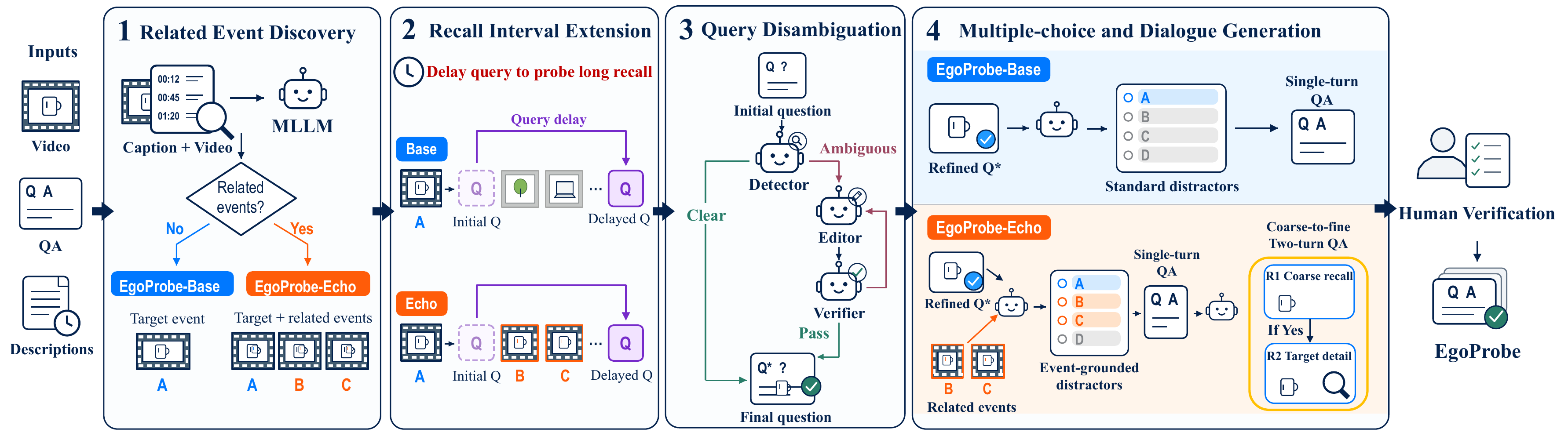}
\caption{\textbf{EgoProbe data engine.} Four automatic stages transform source QA into streaming memory probes: related-event discovery, recall interval extension, query disambiguation, and multiple-choice and dialogue generation. The resulting QA items are then reviewed by human annotators.}
\label{fig:construction_pipeline}
\end{figure}

\subsection{Data Engine for Benchmark Construction}
\label{sec:construction}
\fyq{
\textbf{Overview.}
Our data engine starts from egocentric videos with dense temporal descriptions and existing QA annotations, using QaEgo4D ~\cite{barmann2022did}, which provides QA annotations for Ego4D ~\cite{grauman2022ego4d} videos, and EgoLife ~\cite{yang2025egolife} as source datasets.
Given a video, a target-event QA pair $(q,a)$, timestamped dense captions, and its original evidence interval, we construct two complementary subsets.
\textbf{EgoProbe-Base} contains samples without related events within the recall interval, whereas \textbf{EgoProbe-Echo} contains samples with related events that provide alternative details for constructing event-grounded distractors.
Importantly, both subsets probe long-range recall and are instantiated as single-turn QA, while EgoProbe-Echo further challenges models with event-specific recall and coarse-to-fine multi-turn QA.}

\fyq{The overall construction pipeline summarized in Fig.~\ref{fig:construction_pipeline} illustrates how our data engine transforms source QAs into probes of long-range memory and event-specific recall. Starting from the source QA and its video and timestamped descriptions, the engine progressively transforms each sample through four automatic stages.
1) It first performs \textit{Related-event Discovery} to identify whether the video contains events related to the target event, which determines whether the sample follows the EgoProbe-Base or EgoProbe-Echo track.
2) It then performs \textit{Recall Interval Extension} by delaying the query time to increase the temporal distance from the target event.
3) \textit{Query Disambiguation} further checks and refines the question to ensure that it still uniquely refers to the original target event while preserving the original answer.
4) \textit{Multiple-choice and Dialogue Generation} converts the refined QA into a four-option multiple-choice question; for EgoProbe-Echo samples, this stage additionally introduces event-grounded distractors and derives coarse-to-fine two-turn dialogues.
Finally, \textit{Human Verification} checks the generated QA and associated video evidence to ensure the accuracy and quality of the final benchmark. Note that all resulting probes are defined under a streaming protocol: models process video observations chronologically and maintain memory for subsequent recall, and the query is revealed only at its designated arrival time $t_q$. 
}


\fyq{\textbf{Step 1: Related Event Discovery.}}
\fyq{
To identify recurring event patterns, we first use lexical retrieval to search the timestamped video descriptions after the target event $e^\star$ for other events sharing question-relevant objects, actions, or activities.
An LLM-based module then verifies whether each candidate is a distinct related event and identifies its associated answer. If this answer differs from the target answer, the candidate is retained as a valid contrastive event, and its answer serves as an event-grounded distractor. We retain event timestamps to guide recall-interval extension in Step 2. In Step 4, alternative answers provide event-grounded distractors, while shared event patterns guide coarse-grained question generation.
}

\fyq{\textbf{Step 2:  Recall Interval Extension.}}
\fyq{
We extend the recall interval $\Delta t$ by delaying query arrival according to the video's duration and event structure.
For samples with adjustable query times, we select $t_q$ from four predefined recall-interval ranges: short ($<3$ min), medium ($3$--$30$ min), long ($30$ min--$12$ h), and ultra-long ($\geq 12$ h), subject to video boundaries. The timestamps of related events identified in Step 1 further guide query placement, determining which events fall within the resulting recall interval.
This allows us to control both the temporal distance from the target event and the related events encountered by the model before answering.
}

\noindent\textbf{Step 3: Query Disambiguation.}
\fyq{
Extending the recall interval may introduce ambiguity, as multiple events in the observed history could match the original question. To resolve this, we employ a feedback-guided disambiguation loop with three LLM-based roles: a \textit{detector}, an \textit{editor}, and a \textit{verifier}. The detector first checks whether the question uniquely identifies the target event among all events before $t_q$. If ambiguity is detected, the editor refines the question by adding minimal cues grounded in the target-event evidence while preserving the original answer. The verifier then checks target-event identification, cue grounding, and answer leakage. Questions passing verification are retained; otherwise, the verifier's feedback guides refinement, and unresolved cases are discarded.
}

\fyq{
\textbf{Step 4: Multiple-choice and Dialogue Generation.} Given the refined question from Step 3 and its original answer, we convert the source open-ended QA into a four-option multiple-choice question to enable more consistent evaluation and better leverage the event-grounded distractors identified in Step 1.
For EgoProbe-Base, an LLM generates three plausible incorrect options.
For EgoProbe-Echo, we prioritize event-grounded distractors as hard negatives and use an LLM to generate additional incorrect options when needed.
Each question contains one ground-truth answer and three wrong options, yielding single-turn VQA instances for both tracks.}

\fyq{For EgoProbe-Echo, we additionally derive coarse-to-fine two-turn dialogues to probe different levels of memory recall.
We use the semantic pattern shared by the target and related events to formulate the first turn R1 as a yes/no question with an affirmative ground-truth answer.
The second turn R2 adapts the corresponding single-turn question into a follow-up that retains its target-event cues, correct answer, and four options, requiring the model to retrieve the detail associated with the designated event rather than a competing detail from another observed event.
A separate verification call checks shared-event support, answer leakage, and preservation of the original detail question before accepting the dialogue.
Further construction details are provided in Appendix~\ref{app:construction}.
}

\fyq{\textbf{Human Verification.}}
Trained human annotators review every automatically constructed QA and its associated video evidence through a dedicated web interface. They verify question clarity, target-event identification, answer support and uniqueness, event--distractor associations, and consistency between single-turn questions and derived dialogues. More details are provided in Appendix~\ref{app:validation}.

\begin{figure}[!b]
\centering
\includegraphics[width=\linewidth,trim=0 30bp 0 0,clip]{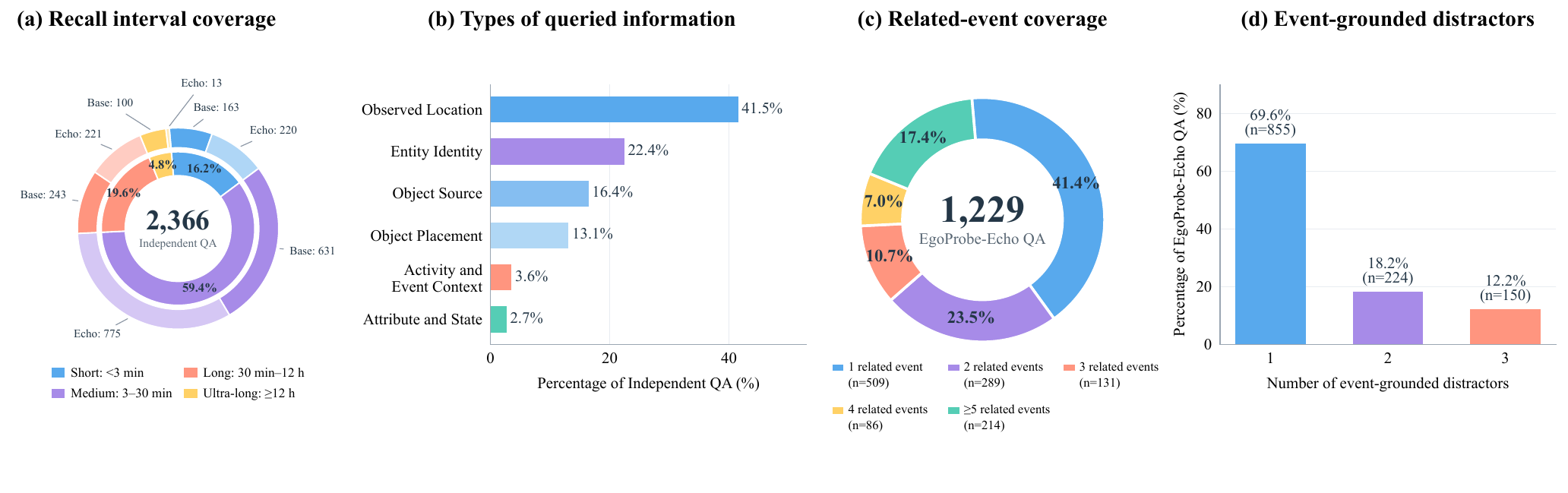}
\setlength{\abovecaptionskip}{2pt}
\caption{\textbf{EgoProbe statistics}: (a) recall intervals and (b) queried information types across EgoProbe; (c) related events and (d) event-grounded distractors per question across EgoProbe-Echo.}
\label{fig:dataset_statistics}
\end{figure}

\subsection{Statistics and Coverage}
\label{sec:statistics}
EgoProbe-Base contains 1,137 single-turn QA pairs, and EgoProbe-Echo contains 1,229 single-turn QA pairs with 1,229 corresponding coarse-to-fine two-turn dialogues, yielding 2,366 independent QA pairs across 671 egocentric videos. Figure~\ref{fig:dataset_statistics} summarizes the benchmark's temporal coverage, queried information types, related-event distribution, and distractor composition.
We divide recall intervals into four categories: Short ($<3$ min), Medium (3--30 min), Long (30 min--12 h), and Ultra-long ($\geq 12$ h). The intervals range from 31.63 seconds to 138.54 hours, with 24.4\% of QA pairs requiring recall after at least 30 minutes. Questions cover location, entity identity, object source and placement, activity and event context, and attributes and states. Within EgoProbe-Echo, 58.6\% of QA pairs have multiple annotated related events, and 30.4\% contain two or three event-grounded distractors. These distributions cover different recall intervals and related-event configurations, supporting evaluation of long-range memory and event-specific recall. Representative QA examples from both subsets are provided in Appendix~\ref{app:examples}.

\begin{table}[!t]
\centering
\definecolor{offlineModel}{HTML}{F2F2F2}
\definecolor{streamWindow}{HTML}{FFF1E6}
\definecolor{streamKV}{HTML}{EDF4FC}
\definecolor{streamEvent}{HTML}{EDF7F0}
\definecolor{streamObject}{HTML}{FFF9E6}
\definecolor{streamReason}{HTML}{F4EFF9}
\setlength{\abovecaptionskip}{0pt}
\caption{Overall accuracy (\%) on EgoProbe. Bold indicates best within each model category. Row colors indicate offline evaluation (\protect\textcolor{offlineModel}{\protect\rule{0.8em}{0.8em}}) and streaming approaches based on recent windows (\protect\textcolor{streamWindow}{\protect\rule{0.8em}{0.8em}}), KV-cache management (\protect\textcolor{streamKV}{\protect\rule{0.8em}{0.8em}}), event-centric memory (\protect\textcolor{streamEvent}{\protect\rule{0.8em}{0.8em}}), object-centric memory (\protect\textcolor{streamObject}{\protect\rule{0.8em}{0.8em}}), or streaming reasoning and interaction (\protect\textcolor{streamReason}{\protect\rule{0.8em}{0.8em}}).}
\label{tab:overall_performance}
\small
\setlength{\tabcolsep}{3pt}
\renewcommand{\arraystretch}{1.15}
\begin{tabularx}{\linewidth}{@{}>{\raggedright\arraybackslash}p{0.24\linewidth}|>{\raggedright\arraybackslash}p{0.23\linewidth}|>{\centering\arraybackslash}p{0.145\linewidth}*{3}{|Y}@{}}
\toprule
\multicolumn{2}{c}{} & \multicolumn{1}{c}{\textbf{EgoProbe-Base}} & \multicolumn{3}{c}{\textbf{EgoProbe-Echo}} \\
\cmidrule(lr){3-3}\cmidrule(l){4-6}
\multicolumn{1}{@{}l}{\textbf{Model}} & \multicolumn{1}{l}{\textbf{Backbone}} & \multicolumn{1}{c}{\textbf{Single-turn}} & \multicolumn{2}{c}{\textbf{Multi-turn}} & \multicolumn{1}{c@{}}{\shortstack{\textbf{Single-turn}}} \\
\cmidrule(lr){4-5}
\multicolumn{2}{c}{} & \multicolumn{1}{c}{\textbf{Acc.}$\uparrow$} & \multicolumn{1}{c}{\textbf{R1 Acc.}$\uparrow$} & \multicolumn{1}{c}{\textbf{R2 Acc.}$\uparrow$} & \multicolumn{1}{c@{}}{\textbf{Acc.}$\uparrow$} \\
\midrule
\multicolumn{6}{c}{\textit{Proprietary MLLMs}} \\
\rowcolor{offlineModel}
Gemini-3.5-Flash & -- & 51.10 & 79.25 & 33.77 & \textbf{43.69} \\
\rowcolor{offlineModel}
GPT-5.6-Sol & -- & \textbf{55.32} & \textbf{89.75} & \textbf{43.69} & 43.37 \\
\midrule
\multicolumn{6}{c}{\textit{Open-source General-purpose MLLMs}} \\
\rowcolor{offlineModel}
Qwen3.5 & Qwen3.5-9B & 45.91 & 81.53 & 34.83 & 35.31 \\
\rowcolor{offlineModel}
InternVL3.5 & Qwen3-8B & 47.58 & 86.66 & 30.43 & 34.91 \\
\rowcolor{offlineModel}
LLaVA-OneVision-2 & Qwen3-8B & \textbf{52.07} & \textbf{93.82} & \textbf{38.24} & \textbf{42.96} \\
\midrule
\multicolumn{6}{c}{\textit{Open-source Long-video Models}} \\
\rowcolor{offlineModel}
Ego-R1 & Qwen2.5-3B-Instruct & 40.99 & 43.53 & 13.10 & 32.14 \\
\rowcolor{offlineModel}
EgoRAG & Qwen3-32B & 36.50 & 44.18 & 15.62 & 31.57 \\
\rowcolor{offlineModel}
VideoChat-Flash & Qwen2-7B & \textbf{53.21} & 92.68 & 38.65 & 41.66 \\
\rowcolor{offlineModel}
EM\textsuperscript{2}Mem & Qwen3.5-35B-A3B & 47.93 & \textbf{93.65} & \textbf{42.72} & \textbf{44.10} \\
\midrule
\multicolumn{6}{c}{\textit{Streaming Models}} \\
\rowcolor{streamWindow}
SimpleStream & Qwen3-VL-8B & 28.41 & 25.71 & 6.75 & 27.42 \\
\rowcolor{streamKV}
StreamingVLM & Qwen2.5-VL-7B & 44.50 & 41.09 & 12.69 & 31.24 \\
\rowcolor{streamKV}
HERMES & Qwen2.5-VL-7B & 46.44 & 61.19 & 21.56 & 34.17 \\
\rowcolor{streamEvent}
StreamForest & Qwen2-7B & \textbf{50.66} & 88.85 & \textbf{35.39} & \textbf{38.89} \\
\rowcolor{streamEvent}
OASIS & Qwen3-VL-8B & 47.41 & 70.55 & 25.06 & 36.21 \\
\rowcolor{streamEvent}
EventMemAgent & Qwen3-VL-8B & 49.52 & 88.36 & 32.63 & 37.75 \\
\rowcolor{streamObject}
MEMO & Qwen3.5-9B & 49.78 & 73.96 & 25.39 & 34.09 \\
\rowcolor{streamObject}
ObjectStream & Qwen2.5-VL-7B & 46.53 & 69.49 & 22.21 & 32.38 \\
\rowcolor{streamReason}
VST & Qwen2.5-VL-7B & 48.64 & \textbf{95.28} & 33.93 & 35.48 \\
\rowcolor{streamReason}
JoyAI-VL-Interaction & JoyAI-VL-1.0 (8B) & 45.29 & 81.86 & 26.53 & 31.49 \\
\bottomrule
\end{tabularx}
\end{table}

\section{Experiments}
\label{sec:experiments}

\subsection{Experimental Setup}
\label{sec:setup}
\paragraph{Evaluated Models.}
We evaluate 19 models across four categories: (1) \emph{proprietary MLLMs}, including Gemini-3.5-Flash~\cite{google2026gemini35flash} and GPT-5.6-Sol~\cite{openai2026gpt56sol}; (2) \emph{open-source general-purpose MLLMs}, including Qwen3.5~\cite{team2025qwen3}, InternVL3.5~\cite{wang2025internvl3}, and LLaVA-OneVision-2~\cite{li2024llava}; (3) \emph{long-video models}, including Ego-R1~\cite{tian2025ego}, EgoRAG~\cite{yang2025egolife}, VideoChat-Flash~\cite{li2026videochat}, and EM\textsuperscript{2}Mem~\mbox{\cite{chen20262mem}};
\fyq{and (4) \emph{streaming models}, covering representative approaches with different design emphases:
recent-frame windows (SimpleStream~\cite{shen2026simple}),
KV-cache management (StreamingVLM~\cite{xu2026streamingvlm} and HERMES~\cite{zhang2026hermes}),
event-centric memory (StreamForest~\cite{zeng2026streamforest}, OASIS~\cite{liang2026oasis}, and EventMemAgent~\cite{wen2026eventmemagent}),
object-centric memory (MEMO~\cite{li2026memo} and ObjectStream~\cite{dong2026objectstream}),
and streaming reasoning and interaction (VST~\cite{guan2026video} and JoyAI-VL-Interaction~\cite{yao2026joyai}).
Model-specific configurations are detailed in Appendix~\ref{app:reproducibility}.}

\paragraph{Evaluation Protocols.} \label{sec:streaming_protocol} \fyq{All models are restricted to video observations available up to query time $t_q$.
Under the \emph{strict streaming} protocol, models process observations continuously without knowing the query and answer using their retained memory state.
Long-video models instead adopt \emph{offline prefix processing}, applying their native encoding or retrieval procedures to the available video history, while general-purpose and proprietary MLLMs receive frames sampled from this history within their input budgets.
For two-turn dialogues, both rounds share the same video evidence, with no additional observations introduced before R2.
}

\paragraph{Evaluation Metrics.}
\label{sec:metrics}

\fyq{Accuracy (\%) is used as primary metric. Single-turn accuracy is the proportion of correctly answered questions. For two-turn dialogues, critically, R2 is counted as correct only when both rounds are answered correctly.}

\subsection{Overall Performance}
\label{sec:overall}
Table~\ref{tab:overall_performance} compares model families on long-range recall in EgoProbe-Base and event-specific recall in EgoProbe-Echo under their respective evaluation protocols.

\noindent\textbf{Overall difficulty.}
EgoProbe remains challenging across model families. Uniform random guessing yields 25\% single-turn accuracy and 12.5\% two-turn completion accuracy when guessing independently across both rounds. Even the best model, GPT-5.6-Sol, achieves only 55.32\% Base accuracy and 43.69\% two-turn completion accuracy. Notably, SimpleStream achieves only 6.75\% two-turn accuracy despite its strong results on OVO-Bench and StreamingBench, highlighting the limitations of recent-frame input and supporting the need for long-range memory.

\noindent\textbf{Comparison across model families.}
Proprietary MLLMs perform strongly overall, while open-source general-purpose models remain competitive: LLaVA-OneVision-2 reaches 52.07\% on EgoProbe-Base. Among long-video models, VideoChat-Flash achieves 53.21\% Base accuracy, and EM\textsuperscript{2}Mem reaches 42.72\% two-turn accuracy. These results use query-time access to historical video, unlike memory maintained during streaming observation. Among streaming models, StreamForest leads on both tasks with 50.66\% and 35.39\%, respectively. The remaining gap highlights the difficulty of supporting reliable recall under streaming constraints.

\noindent\textbf{Recall among related experiences.}
All models obtain lower single-turn accuracy on EgoProbe-Echo than on EgoProbe-Base. Although the subsets are not matched controls, this consistent pattern indicates greater difficulty in answering about target-event details among related experiences.

\noindent\textbf{Coarse versus precise recall.}
The coarse-to-fine results further expose this challenge: VST achieves 95.28\% R1 accuracy but only 33.93\% two-turn completion accuracy, showing that a correct coarse response does not ensure accurate target-detail recall.

\noindent\textbf{Single-turn versus two-turn performance.}
Echo single-turn accuracy exceeds two-turn completion accuracy for most models, indicating that success on a standalone detail question does not necessarily carry over to the two-turn task. This comparison motivates examining models' ability to answer consistently across turns, while also accounting for failures at the initial R1 stage.

\FloatBarrier
\begingroup
\setlength{\intextsep}{4pt}
\setlength{\textfloatsep}{4pt}
\begin{table}[!htbp]
\centering
\begin{minipage}[t]{0.50\linewidth}
\vspace{0pt}
\caption{EgoProbe-Base accuracy (\%) across recall intervals. Bold marks column maxima.}
\label{tab:temporal_gap}
\scriptsize
\setlength{\tabcolsep}{1.5pt}
\renewcommand{\arraystretch}{1.2}
\begin{tabularx}{\linewidth}{@{}l*{4}{Y}@{}}
\toprule
\textbf{Model} & \shortstack{\textbf{Short}\\\textbf{$<3$ min}} & \shortstack{\textbf{Medium}\\\textbf{3--30 min}} & \shortstack{\textbf{Long}\\\textbf{30 min--12 h}} & \shortstack{\textbf{Ultra-long}\\\textbf{$\geq 12$ h}} \\
\midrule
StreamingVLM & 45.40 & 45.17 & 47.74 & 31.00 \\
HERMES & 46.01 & 50.40 & 44.03 & 28.00 \\
StreamForest & 50.92 & 53.72 & \textbf{48.15} & 37.00 \\
OASIS & 52.15 & 50.55 & 42.39 & 32.00 \\
EventMemAgent & 46.63 & 52.93 & \textbf{48.15} & 36.00 \\
MEMO & 46.01 & \textbf{54.68} & 43.62 & \textbf{40.00} \\
ObjectStream & 44.79 & 49.76 & 43.62 & 36.00 \\
VST & \textbf{53.99} & 50.55 & 44.03 & 39.00 \\
JoyAI-VL-Interaction & 48.47 & 48.97 & 38.68 & 33.00 \\
\bottomrule
\end{tabularx}
\end{minipage}\hspace{0.02\linewidth}%
\begin{minipage}[t]{0.48\linewidth}
\vspace{0pt}
\centering
\makeatletter\def\@captype{figure}\makeatother
\caption{Two-turn outcomes on EgoProbe-Echo (\% of all dialogues).}
\label{fig:echo_multiturn}
\includegraphics[width=\linewidth]{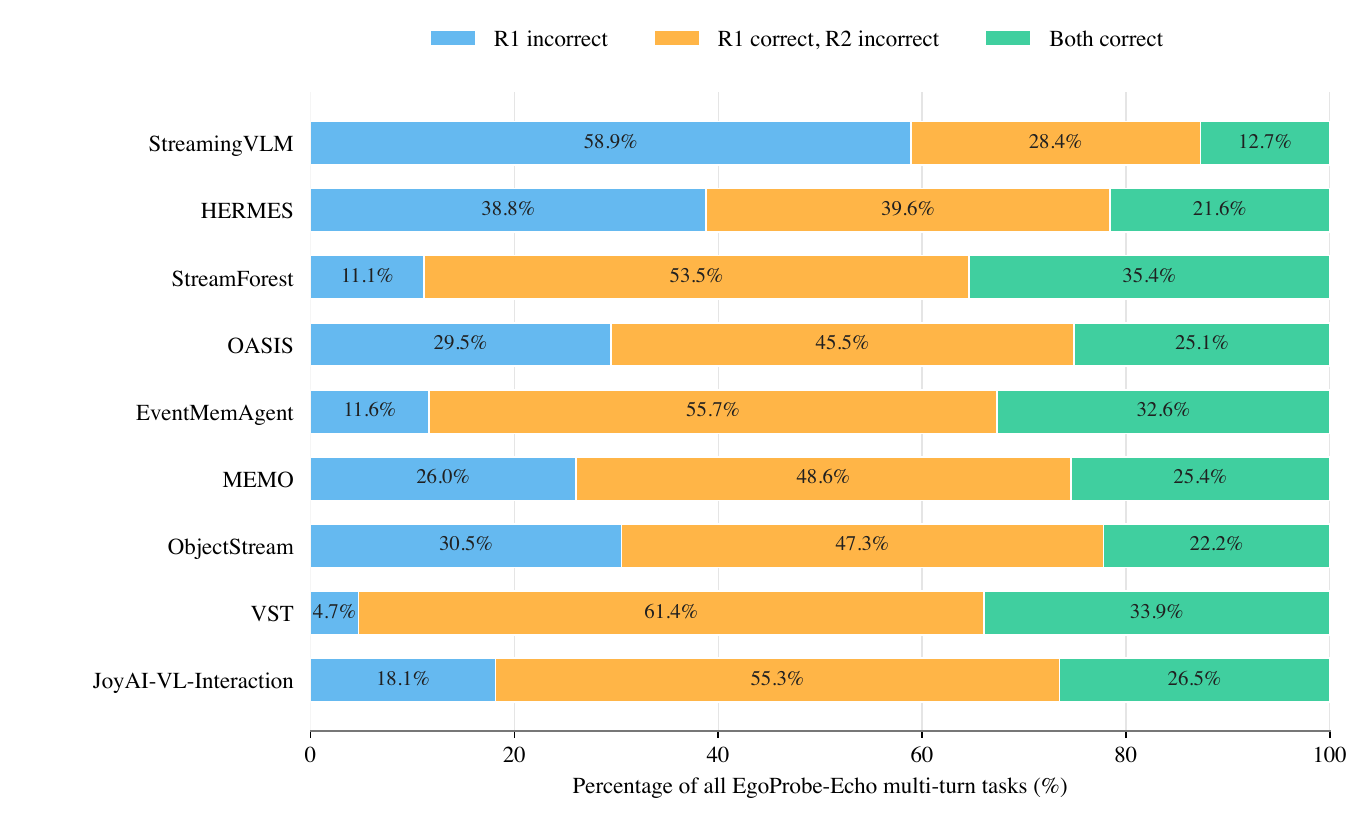}
\end{minipage}
\end{table}



\FloatBarrier
\endgroup

\subsection{Long-Range Recall}
\label{sec:temporal_gaps}
We examine recall intervals on EgoProbe-Base, whose verified intervals contain no intervening related events. This separates recall interval from related-event interference, allowing us to examine how performance varies with the time elapsed since the target event. Table~\ref{tab:temporal_gap} reports streaming-model accuracy across the four recall-interval groups defined in Section~\ref{sec:statistics}. Every model attains its lowest accuracy in the Ultra-long group, where scores range from 28.00\% to 40.00\%. These results highlight the difficulty of recalling distant target events even without intervening related events. Figure~\ref{fig:base_recall_interval} in Appendix~\ref{app:results_3} visualizes accuracy across recall intervals.

\subsection{Event-Specific Recall under Interference}
\label{sec:occurrence_errors}
We use EgoProbe-Echo single-turn QA to analyze whether models recall the target-event answer, select a detail from another related experience, or select another incorrect option. Table~\ref{tab:occurrence_errors} groups questions by one, two, or at least three related events supporting alternative answers. Alongside accuracy, we report the \emph{event-linked error rate} and \emph{other-option error rate}, defined as the proportions of all questions answered with event-grounded and other incorrect options, respectively. The random-choice row accounts for the distribution of these option types within each group.
\begin{table}[tbp]
\centering
\setlength{\abovecaptionskip}{0pt}
\caption{Single-turn answer outcomes (\%) on EgoProbe-Echo. Event err. and Other err. denote event-linked and other-option error rates, respectively. Bold marks the highest accuracy and lowest error rates among models within each event group. The gray row denotes random choice.}
\label{tab:occurrence_errors}
\scriptsize
\setlength{\tabcolsep}{0.8pt}
\renewcommand{\arraystretch}{1.15}
\begin{tabularx}{\linewidth}{@{}l*{9}{Y}@{}}
\toprule
& \multicolumn{3}{c}{\textbf{1 event}} & \multicolumn{3}{c}{\textbf{2 events}} & \multicolumn{3}{c}{$\geq$\textbf{3 events}} \\
\cmidrule(lr){2-4}\cmidrule(lr){5-7}\cmidrule(l){8-10}
\textbf{Model} & \textbf{Acc.}$\uparrow$ & \textbf{Event err.}$\downarrow$ & \textbf{Other err.}$\downarrow$ & \textbf{Acc.}$\uparrow$ & \textbf{Event err.}$\downarrow$ & \textbf{Other err.}$\downarrow$ & \textbf{Acc.}$\uparrow$ & \textbf{Event err.}$\downarrow$ & \textbf{Other err.}$\downarrow$ \\
\midrule
\rowcolor{black!5}
Random choice & 25.00 & 25.00 & 50.00 & 25.00 & 36.93 & 38.07 & 25.00 & 53.01 & 21.99 \\
\midrule
StreamingVLM & 37.08 & 35.47 & 27.45 & 22.41 & 52.28 & 25.31 & 27.12 & 63.56 & 9.32 \\
HERMES & 36.44 & 36.44 & 27.13 & 31.54 & 46.89 & 21.58 & \textbf{32.05} & \textbf{60.55} & 7.40 \\
StreamForest & 43.18 & 38.04 & \textbf{18.78} & \textbf{39.83} & \textbf{44.40} & 15.77 & 30.96 & 61.64 & 7.40 \\
OASIS & 42.54 & 34.19 & 23.27 & 29.88 & 56.02 & \textbf{14.11} & 29.59 & 65.21 & \textbf{5.21} \\
EventMemAgent & \textbf{44.46} & \textbf{31.94} & 23.60 & 33.61 & 46.89 & 19.50 & 29.04 & 63.84 & 7.12 \\
MEMO & 40.29 & 37.08 & 22.63 & 27.80 & 56.43 & 15.77 & 27.67 & 64.66 & 7.67 \\
ObjectStream & 36.44 & 37.40 & 26.16 & 30.71 & 52.28 & 17.01 & 26.58 & 66.58 & 6.85 \\
VST & 42.38 & 37.72 & 19.90 & 30.29 & 53.11 & 16.60 & 27.12 & 66.85 & 6.03 \\
JoyAI-VL-Interaction & 36.28 & 36.76 & 26.97 & 29.05 & 51.87 & 19.09 & 24.93 & 65.21 & 9.86 \\
\bottomrule
\end{tabularx}
\end{table}

\paragraph{Related events challenge precise recall and reveal event-linked errors.}
All models achieve lower accuracy in multiple-event groups than in the single-event group. For example, StreamForest's accuracy decreases from 43.18\% with one related event to 39.83\% with two and 30.96\% with at least three. Meanwhile, event-linked error rates increase across the three groups for every model and consistently exceed their random-choice references. These results highlight difficulty of recalling the correct detail when similar experiences contain different answers. EgoProbe captures this challenge by identifying answers that are valid for another event but incorrect for the queried event.

\paragraph{Coarse-to-fine probing distinguishes failure levels.}
While event links characterize the details selected in incorrect answers, two-turn probing examines the level of recall at which responses fail. Figure~\ref{fig:echo_multiturn} decomposes all EgoProbe-Echo dialogues into three outcomes: incorrect R1, correct R1 followed by incorrect R2, and both rounds correct. For VST, 61.4\% of all dialogues have a correct R1 response but an incorrect R2 response. This outcome also accounts for over half of dialogues for StreamForest, EventMemAgent, JoyAI-VL. Staged probing therefore distinguishes failure to answer the coarse question from failure to recall the target-specific detail after a successful coarse response.

\section{Conclusion and Limitations}
\label{sec:conclusion}
We introduce EgoProbe, a benchmark for long-range memory and event-specific recall in streaming egocentric video. Through an event-grounded data engine, EgoProbe constructs memory probes that evaluate both how far models can recall and how exact they recover target-event details. Our evaluation reveals substantial limitations in both dimensions: models struggle to recall distant events and recover target-specific details. EgoProbe provides a testbed for developing and diagnosing memory capabilities in streaming egocentric systems.
\paragraph{Limitations.}
Our framework supports longer streams, but EgoProbe's temporal coverage is limited by the available EgoLife recordings. Future work will explore recordings spanning weeks or months to evaluate longer-term memory. EgoProbe also omits audio information, an important component of egocentric experiences; incorporating it would enable a more comprehensive evaluation of egocentric memory.

\bibliography{iclr2027_conference}
\bibliographystyle{iclr2027_conference}

\clearpage
\appendix
\section{Benchmark Examples}
\label{app:examples}
The following examples illustrate long-range recall in EgoProbe-Base and event-specific recall in EgoProbe-Echo. Each panel presents selected video frames, followed by the question and answer options. For Echo, the options are shared by the single-turn question and R2. Event-grounded distractors are supported by other related experiences but remain incorrect for the target event; other options have no such event association. 

\noindent\begin{minipage}{\linewidth}
\begingroup
\centering
\includegraphics[width=\linewidth]{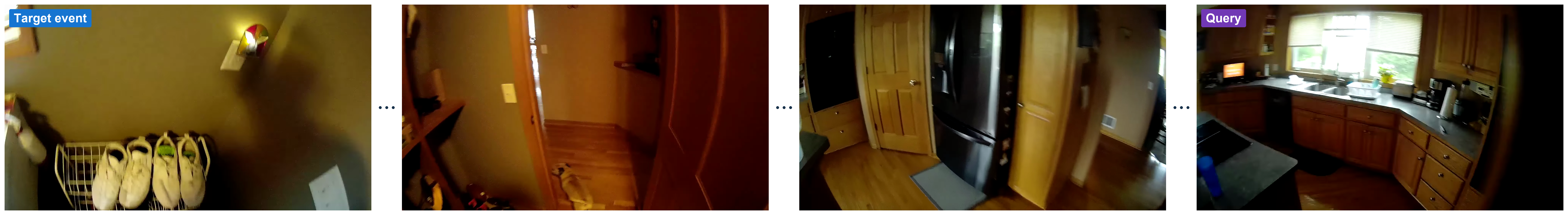}

\makeatletter\def\@captype{figure}\makeatother
\caption{EgoProbe-Base single-turn example. The model must recall the earlier location of the shoes after approximately 17 minutes.}
\label{fig:appendix_base_example}
\par\endgroup
\smallskip
\begin{minipage}{\linewidth}
\small

\textbf{EgoProbe-Base: recalling the location of shoes.}
Target evidence: 12:15--12:17; query time: 29:19; recall interval: 17 min 2 s.

\smallskip
\textbf{Single-turn question:} In what location did I see my shoes?
\par\smallskip
\begin{tabularx}{\linewidth}{@{}lXl@{}}
\toprule
\textbf{Option} & \textbf{Answer} & \textbf{Role}\\
\midrule
A & inside the shoe cabinet & Other option\\
B & under the bed & Other option\\
C & on the floor beside the door & Other option\\
\textbf{D} & \textbf{on the shoe rack} & \textbf{Correct answer}\\
\bottomrule
\end{tabularx}
\par\smallskip
The target evidence shows the shoes on a shoe rack. The question is asked after subsequent observations, with no intervening related event in the verified recall interval. All three incorrect options are other options rather than event-grounded distractors.
\end{minipage}
\end{minipage}
\par\medskip

\noindent\begin{minipage}{\linewidth}
\begingroup
\centering
\includegraphics[width=\linewidth]{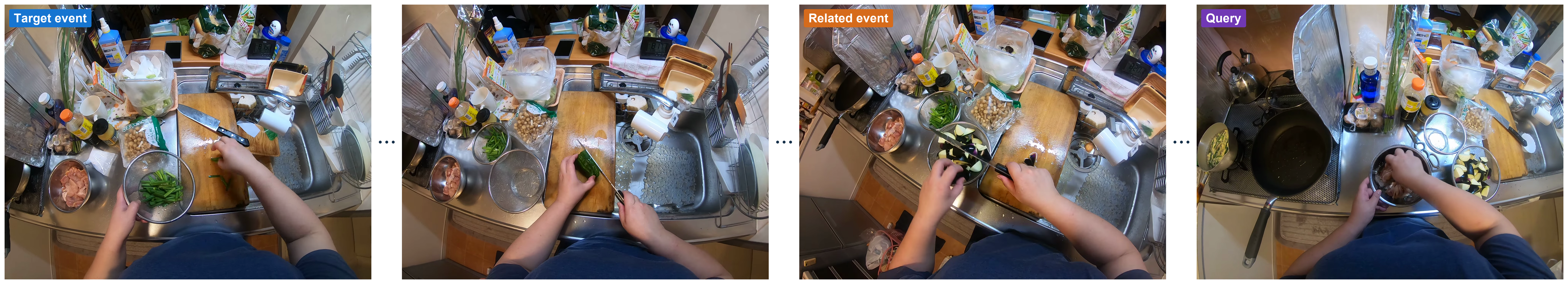}

\makeatletter\def\@captype{figure}\makeatother
\caption{EgoProbe-Echo example with one event-grounded distractor. R1 asks about shared event content, while the single-turn question and R2 require the target-specific ingredient.}
\label{fig:appendix_echo_example}
\par\endgroup
\smallskip
\begin{minipage}{\linewidth}
\small
\textbf{EgoProbe-Echo: recalling food placed in a sieve.}
Target evidence: 02:59.09--03:12.50; query time: 11:17; recall interval: 8 min 4.50 s.

\smallskip
\textbf{Single-turn question:} What did I put in the sieve right after picking it from the dish drainer?

\smallskip
\textbf{Two-turn dialogue.} \textbf{R1:} Did I put food into the sieve? \textbf{Answer: Yes} (choices: Yes / No).
\par\textbf{R2, after Yes:} What did I put in the sieve right after picking it from the dish drainer?
\par\smallskip
\begin{tabularx}{\linewidth}{@{}lXl@{}}
\toprule
\textbf{Option} & \textbf{Answer (single-turn and R2)} & \textbf{Role}\\
\midrule
A & sago & Other option\\
B & cabbage & Other option\\
\textbf{C} & \textbf{spring onion} & \textbf{Correct answer}\\
D & eggplant & Event-grounded distractor\\
\bottomrule
\end{tabularx}
\par\smallskip
The wearer picks up the sieve from the dish drainer at 02:59.35 and puts spring onion into it at 03:00.42. A later event at 07:57.58 involves putting eggplant into the sieve, supporting option D. Both the target and the eggplant event support Yes in R1; the cue ``right after picking it from the dish drainer'' identifies the target for R2.
\end{minipage}
\end{minipage}
\par\medskip

\clearpage
\noindent\begin{minipage}{\linewidth}
\begingroup
\centering
\includegraphics[width=\linewidth]{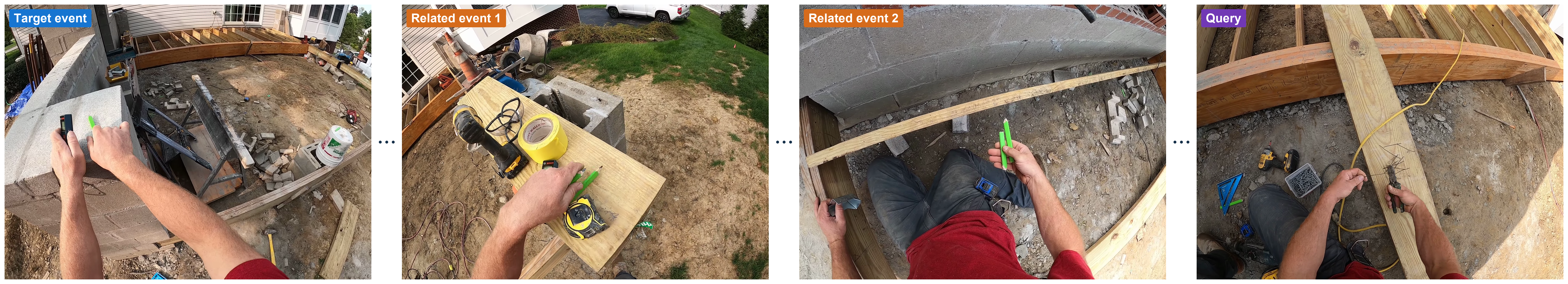}

\makeatletter\def\@captype{figure}\makeatother
\caption{EgoProbe-Echo example with two event-grounded distractors. The wooden-plank and pocket locations are supported by later related events but are incorrect for the event preceding use of the gauge.}
\label{fig:appendix_echo_pencil}
\par\endgroup
\smallskip
\begin{minipage}{\linewidth}
\small
\textbf{EgoProbe-Echo: recalling where a green pencil was placed.}
Target evidence: 27:16.95--27:19.16; query time: 01:06:07; recall interval: 38 min 47.84 s.

\smallskip
\textbf{Single-turn question:} Where did I put the green pencil before using the gauge?

\smallskip
\textbf{Two-turn dialogue.} \textbf{R1:} Did I place a pencil? \textbf{Answer: Yes} (choices: Yes / No).
\par\textbf{R2, after Yes:} Where did I put the green one before using the gauge?
\par\smallskip
\begin{tabularx}{\linewidth}{@{}lXl@{}}
\toprule
\textbf{Option} & \textbf{Answer (single-turn and R2)} & \textbf{Role}\\
\midrule
A & in my pocket & Event-grounded distractor\\
\textbf{B} & \textbf{on the wall} & \textbf{Correct answer}\\
C & on the wooden plank & Event-grounded distractor\\
D & on the table & Other option\\
\bottomrule
\end{tabularx}
\par\smallskip
The target event places the pencil on the wall before the wearer uses the gauge at 27:20.00. Later, the pencil is placed on a wooden plank at 29:45.62 and in a pocket at 50:06.03, supporting options C and A, respectively. These experiences all support Yes in R1, but only the wall location answers the target-specific question. R2 uses ``the green one'' to refer back to the pencil in R1. For consistent first-person wording in this illustration, option A is displayed as ``in my pocket'' instead of the source wording ``in his pocket.''
\end{minipage}
\end{minipage}
\par\medskip

\section{Detailed Benchmark Construction}
\label{app:construction}
This section expands the four stages in Section~\ref{sec:construction}. We preserve source QA and target evidence as construction anchors, while recording query-time decisions, question revisions, and event--option associations. Human verification is described separately in Appendix~\ref{app:validation}.

\subsection{Source Data}
\label{app:construction_1}
For QaEgo4D, we use QA aligned with Ego4D temporal evidence and timestamped narrations. For EgoLife, we use episodic QA and dense video descriptions from the A1\_JAKE stream. EgoLife questions marked as requiring audio are excluded from the visual-only construction pool. The English description index preserves the original caption identifiers and timestamps, allowing retrieved descriptions to be traced back to video evidence. We use \emph{video descriptions} for both sources throughout the paper. Each construction record contains the source question, canonical answer, target evidence, and stream identity. Target evidence may comprise multiple segments; the latest evidence endpoint determines the recall interval. 

\subsection{Related-Event Discovery}
\label{app:construction_2}
We retrieve descriptions using question-relevant entities, actions, and relations, with lexical matching and compatible aliases. QaEgo4D searches after the target evidence within the source video; EgoLife searches between the target evidence and the fixed native query. Nearby descriptions are grouped into candidate events rather than counted independently. In our setting, the QaEgo4D configuration groups descriptions separated by at most 8 seconds into windows of at most 30 seconds.

An LLM then assesses whether each candidate describes a distinct related experience and whether its evidence supports an answer to the requested detail. Related events with the same answer are retained as context but do not supply alternative answer options. Overlapping descriptions of the same experience are consolidated, retaining their supporting evidence. Each retained event therefore has a temporal reference, a relationship to the target, and an answer association when supported. Event counts refer to these annotated, deduplicated experiences.

\subsection{Recall-Interval Extension and Sample Selection}
\label{app:construction_3}
Query-time planning selects a legal time after the target and the included related-event evidence, bounded by the video endpoint and, where applicable, the next related event outside the selected context. Sampling balances temporal coverage and single- versus multiple-related-event conditions, with a per-batch cap of five targets from one video. 

\subsection{Question Disambiguation}
\label{app:construction_4}
For adjustable questions, a detector evaluates whether the original wording uniquely identifies the target among the events preceding the planned query. Clear questions bypass rewriting. For ambiguous questions, an editor introduces only necessary cues supported by the target's local evidence, preserving the requested fact and answer. A verifier checks the revised reference, cue support, and answer leakage. Failed checks are returned to the editor, with at most two rewrite attempts; unresolved candidates are excluded.

Two retained examples illustrate how evidence-supported cues resolve event ambiguity. Both have complete detection, rewriting, and verification records.

\medskip
\noindent\begin{minipage}{\linewidth}
\small
\textbf{Example 1: Disambiguation through an action cue}
\par\smallskip
\renewcommand{\arraystretch}{1.15}
\begin{tabularx}{\linewidth}{@{}>{\raggedright\arraybackslash}p{0.32\linewidth}>{\raggedright\arraybackslash}X@{}}
\toprule
\textbf{Original question} & Where was the spoon before I picked it?\\
\textbf{Refined question} & Where was the spoon before I picked it \textbf{and cleaned it with a tissue}?\\
\midrule
\textbf{Ambiguity} & The wearer picks up a spoon from a plate (47:29.97), a baking table (48:45.20), and a tray (52:04.10).\\
\textbf{Grounded cue} & Cleaning the spoon with a tissue (47:31.94) identifies the target occurrence without revealing its prior location.\\
\midrule
\textbf{Correct answer} & A. in a plate (unchanged)\\
\textbf{Event-grounded distractors} & C. on a tray; D. on the baking table\\
\textbf{Other option} & B. in the cabinet\\
\bottomrule
\end{tabularx}
\end{minipage}

\medskip
\noindent\begin{minipage}{\linewidth}
\small
\textbf{Example 2: Disambiguation through a location cue}
\par\smallskip
\renewcommand{\arraystretch}{1.15}
\begin{tabularx}{\linewidth}{@{}>{\raggedright\arraybackslash}p{0.32\linewidth}>{\raggedright\arraybackslash}X@{}}
\toprule
\textbf{Original question} & What hand tool did I pick?\\
\textbf{Refined question} & What hand tool did I pick \textbf{from the wall}?\\
\midrule
\textbf{Ambiguity} & The wearer picks up an air pump from a wall (18:09.16), a wrench from a machine (20:23.66), and a putty knife from a lift stand (21:01.87).\\
\textbf{Grounded cue} & The wall location identifies the target occurrence while leaving the tool's identity to be recalled.\\
\midrule
\textbf{Correct answer} & D. air pump (unchanged)\\
\textbf{Event-grounded distractors} & B. wrench; C. putty knife\\
\textbf{Other option} & A. pliers\\
\bottomrule
\end{tabularx}
\end{minipage}

\subsection{Multiple-Choice and Dialogue Generation}
\label{app:construction_5}
Every accepted single-turn task has one correct answer and three incorrect options. For EgoProbe-Base, the construction module generates or selects plausible alternatives of the required answer type without attaching intervening related-event evidence. For EgoProbe-Echo, answers supported by other related events are prioritized as event-grounded distractors, and remaining slots are filled with other options. EgoLife can reuse its original incorrect choices. Option verification checks distinctness, semantic equivalence, answer support, and whether an alternative is still incorrect for the specified target.

Two-turn generation operates on accepted EgoProbe-Echo questions. A shared semantic anchor is extracted from the target and selected related events, and R1 asks whether the corresponding experience occurred. R2 is a conversational adaptation of the single-turn question with its target cues, correct answer, and options preserved. A separate verification call checks support across the selected events, absence of answer leakage, and preservation of the detail task. Parent and dialogue records retain their correspondence and share the same query time and visual evidence.

\subsection{Models, Prompts, and Implementation Settings}
\label{app:construction_6}
The construction stage employs \texttt{gemini-3.7-flash} for all LLM roles. QaEgo4D semantic assessment and dialogue construction use the medium thinking setting; EgoLife single-turn processing uses the low setting and dialogue construction uses medium. The generation limit is 16,384 output tokens for these stages. Structured responses are checked against the expected fields, and verification is performed in a separate call.

Table~\ref{tab:construction_prompt_roles} provides a concise overview of each role's core responsibilities and key prompt instructions, rather than the full prompts. Each call additionally supplies the relevant question and evidence records with stable event identifiers; dynamic sample payloads and output schemas are specified separately from the role instructions.
\begin{table}[htbp]
\centering\small
\caption{Core instructions used by the construction roles.}
\label{tab:construction_prompt_roles}
\begin{tabularx}{\linewidth}{@{}p{0.21\linewidth}X@{}}
\toprule
\textbf{Role} & \textbf{Instruction and constraint}\\
\midrule
Event assessment & Determine whether the candidate is a distinct related event. Extract an answer only when its own descriptions support the queried detail; retrieval similarity is not evidence.\\
Clarity detector & Apply the full question to the supplied pre-query events. Identify alternative plausible referents without answering or rewriting the question.\\
Question editor & Add minimal target-grounded cues that distinguish the target. Preserve the queried fact and correct answer, and do not reveal that answer.\\
Revision verifier & Check target identification, evidence for added cues, and answer leakage. Return specific feedback when a revision fails.\\
Option construction & Prioritize safely supported related-event answers, fill the remaining slots, and reject equivalent options or additional correct answers.\\
Dialogue construction & Identify shared content supported by the selected events, ask a coarse yes/no question, and adapt the original detail question into a follow-up.\\
Dialogue verification & Check shared-event support, coherence, answer leakage, and preservation of the original answer and options.\\
\bottomrule
\end{tabularx}
\end{table}

\section{Human Verification}
\label{app:validation}
\subsection{Review Interface and Evidence}
\label{app:validation_1}
\begin{figure}[htbp]
\centering
\includegraphics[width=\linewidth]{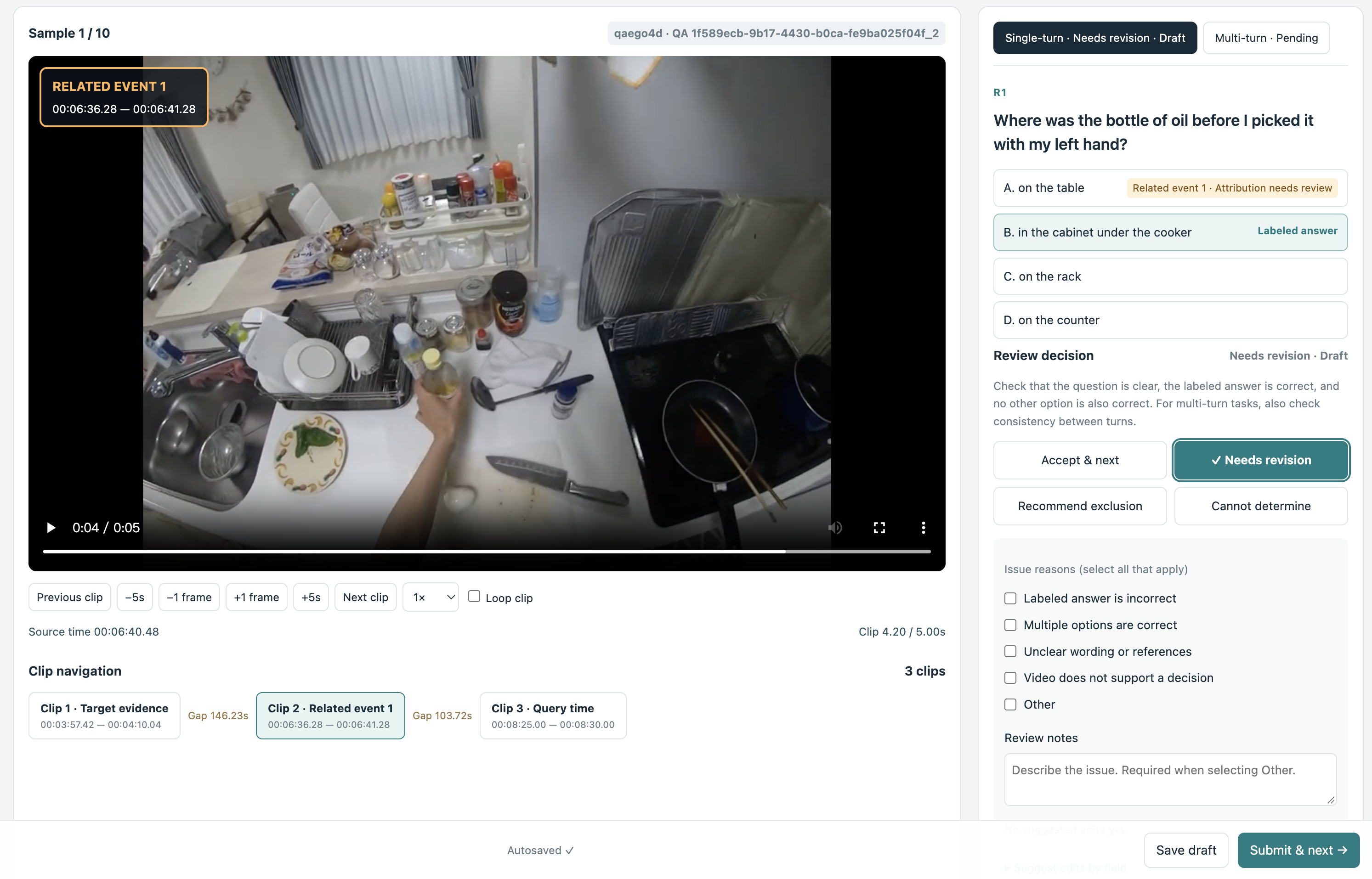}
\caption{Human-review interface. Annotators navigate timestamped evidence clips, check QA and event-linked options, and record decisions with reasons and suggested edits. Labels shown in the interface are subject to verification.}
\label{fig:review_gui}
\end{figure}

Professional annotators use a dedicated web interface to inspect the target, related-event, and query clips together with their source timestamps (Figure~\ref{fig:review_gui}). The interface displays the question, options, original answer label, and event--option links, and supports separate single-turn and multi-turn decisions. Annotators are instructed to assess all supplied evidence rather than assume that the original answer or target/related-event labels are correct.

\subsection{Annotation Guidelines}
\label{app:validation_2}
Review follows six criteria: (1) the question clearly identifies the intended event, entity, and temporal relation; (2) the target video supports the labeled answer; (3) the options are distinct and exactly one answers the question; (4) the derived dialogue preserves the single-turn meaning and remains coherent across rounds; (5) EgoLife person references are understandable from visual evidence, with names replaced by unambiguous descriptions when appropriate; and (6) each event-linked incorrect option is supported by its associated clip while remaining incorrect for the target. Other options do not require supporting related-event clips.

\subsection{Review Procedure}
\label{app:validation_3}
Annotators select \emph{Accepted}, \emph{Needs revision}, \emph{Recommend exclusion}, or \emph{Cannot determine}, and provide reasons and field-level suggestions when needed. A single-turn question and its derived dialogue form one review unit, but their decisions are submitted separately; both dialogue rounds are reviewed together. Suggested edits are kept separate from the original QA until they are applied and checked.

The interface-based review checks QA quality and event correspondence in the supplied clips. Subsequent human verification reviews the full video to confirm the absence of related events within the EgoProbe-Base recall intervals. Claims that require broader temporal context, such as first or last occurrences, are checked against the relevant history before acceptance.

\subsection{Review Coverage and Outcomes}
\label{app:validation_4}
Before human filtering, the candidate pool contains 1,264 EgoProbe-Base single-turn questions and 1,291 EgoProbe-Echo single-turn questions, each of the latter paired with a two-turn dialogue. This gives 2,555 independent QA instances and 3,846 evaluation tasks. 

After human review and filtering, the final benchmark contains 1,137 Base questions, 1,229 Echo questions, and 1,229 corresponding dialogues: 2,366 independent QA instances and 3,595 evaluation tasks. The retained task count is 93.5\% of the pre-filtering pool, with a net reduction of 251 tasks. This is the final retention rate, including items accepted after revision. Annotators record issues and suggested corrections, and revised items are rechecked before final acceptance.

\section{Evaluation Protocols and Implementation Details}
\label{app:reproducibility}
\subsection{History-Access Protocols}
\label{app:reproducibility_1}
All settings enforce the query-time cutoff: model inputs exclude observations after $t_q$, and benchmark target timestamps, answer labels, construction descriptions, and event--option links are withheld from the model. History access nevertheless differs across implementations. To enforce strict streaming evaluation, streaming models process observations chronologically without access to future questions and answer each query using only their retained memory and explicitly stored observations, without reloading discarded video. Prefix-processing references sample or retrieve permitted historical information at query time.

We adapt models where needed to enforce strict streaming evaluation while preserving their original memory mechanisms. For VST, frames are sampled and grouped into fixed-duration chunks independently of query arrival. Each completed chunk updates textual memory through the official thought-generation procedure, after which old frames are released except for a recent-video cache. Query arrival does not trigger an additional persistent memory update. At query time, the model answers using this memory and the available recent frames in an isolated branch. R1 and R2 share dialogue context, but their responses are not written back to the persistent memory, allowing observation to continue independently of the questions. For EventMemAgent, we use incremental event-memory construction: new observations update the retained event memory throughout the stream, and queries are answered from the accumulated state without rebuilding memory from the full video prefix.

\begin{table}[htbp]
\centering\small
\caption{Input and memory settings recovered from evaluation records. Frame budgets describe the available input, not the full duration of the source stream.}
\label{tab:appendix_model_settings}
\setlength{\tabcolsep}{3pt}
\begin{tabularx}{\linewidth}{@{}p{0.25\linewidth}X@{}}
\toprule
\textbf{Model} & \textbf{Input and principal settings}\\
\midrule
Gemini / GPT & 128 prefix frames; maximum side 448 pixels; high image detail.\\
Qwen3.5 & 128 prefix frames; per-frame pixel budget 200,704; non-thinking.\\
InternVL3.5 & 128 prefix frames; one $448\times448$ patch per frame.\\
LLaVA-OneVision-2 & 128 prefix frames; frame input rather than codec input.\\
Ego-R1 & 1 FPS; 30-frame caption chunks; agent limit of seven turns; temporal video retrieval.\\
EgoRAG & 1 FPS; 30-frame caption chunks; Qwen3-32B text retrieval/answering.\\
VideoChat-Flash & Native dynamic sampling, up to 512 prefix frames; 448-pixel model.\\
EM\textsuperscript{2}Mem & Precomputed memory graphs; retrieval limits: 5 episodic, 8 semantic, and 3 visual items; up to three rounds.\\
\midrule
SimpleStream & 1 FPS; a recent window of four frames in the available run configuration.\\
StreamingVLM & 2 FPS; two-frame chunks; visual window 16; text sink and sliding window 512 each.\\
HERMES & 1 FPS; 16-frame chunks; KV budget 6,000.\\
StreamForest & 1 FPS; native online generation and memory update.\\
OASIS & 2 FPS; buffer limit 32 frames; current window 16; up to 16 frames per event node; event retrieval limit 2.\\
EventMemAgent & Incremental event-memory construction at 1 FPS; up to ten assistant turns; Qwen3-VL-4B caption service.\\
MEMO & 1 FPS; entity-aware semantic chunking; top-3 retrieval; 8-frame answer budget; 2,048 cached frames; 512 CPU chunks; 128 output tokens.\\
ObjectStream & 1 FPS; persistent object memory; recent/medium windows of 8/256 frames; 256 entity tracks, 32 tokens per entity; temporal-burst retention; 128 output tokens.\\
VST & 2 FPS; all textual thoughts retained; 5,000 tokens per thought and 16 tokens per answer.\\
JoyAI-VL-Interaction & Journals record chronological observation, eight-frame input chunks, and retained hierarchical summaries.\\
\bottomrule
\end{tabularx}
\end{table}

\subsection{Model Configurations}
\label{app:reproducibility_2}
Table~\ref{tab:appendix_model_settings} summarizes the models run configurations. Model-specific processing budgets are retained rather than equated across architectures. General-purpose visual baselines use 128 sampled prefix frames; VideoChat-Flash uses a larger budget of up to 512 frames. API requests use 128 frames resized to a maximum side of 448 pixels. We evaluate Gemini-3.5-Flash and GPT-5.6-Sol; the API answer-generation limit is 8,192 tokens.

Qwen3.5 uses non-thinking generation with temperature 0.7, top-$p$ 0.8, top-$k$ 20, and a 16,384-token output limit. LLaVA-OneVision-2 uses deterministic decoding with a 1,024-token limit. HERMES uses temperature zero and a 16-token answer limit; StreamingVLM uses option-logit selection for the final answer. 

\subsection{Evaluation Procedure}
\label{app:reproducibility_3}
Predictions are joined to the fixed evaluation lists by sample identifier: 1,137 EgoProbe-Base single-turn questions, 1,229 EgoProbe-Echo single-turn questions, and their 1,229 derived dialogues. The source-question and dialogue correspondence is preserved throughout evaluation.

In the shared streaming harness, queries are processed in chronological order. Each QA is answered in an isolated branch of the current state, preventing one question or its answer from altering the persistent memory used by another. An incomplete observation chunk at the query boundary is handled within that branch without consuming future frames. For a dialogue, the actual R1 response is retained as conversational context for R2. A parsed affirmative response triggers R2; a negative or invalid R1 response does not. Both rounds use the same visual evidence, and only their dialogue context changes.

\subsection{Metrics and Scoring}
\label{app:reproducibility_4}
Single-turn accuracy is the number of correct answers divided by the fixed number of questions. R1 accuracy uses all dialogues as its denominator. R2 accuracy also uses all dialogues and counts a task as correct only when both rounds succeed; dialogues that never reach R2 contribute zero. The released R1 questions are constructed from observed events and have affirmative labels. The 12.5\% two-turn chance reference therefore denotes independent uniform guessing over the two R1 choices and four R2 choices, not an always-affirmative baseline.

For event-linked analysis, let $N$ be the number of questions in a group and $k_i$ the number of distinct event-grounded incorrect options for question $i$. Event-linked error rate counts selections of those options and divides by $N$. Its uniform-choice reference is $100\sum_i(k_i/4)/N$. Other-option errors count selections of the remaining incorrect options.

\begin{figure}[!ht]
\centering
\includegraphics[width=0.8\linewidth,trim=0 12bp 0 0,clip]{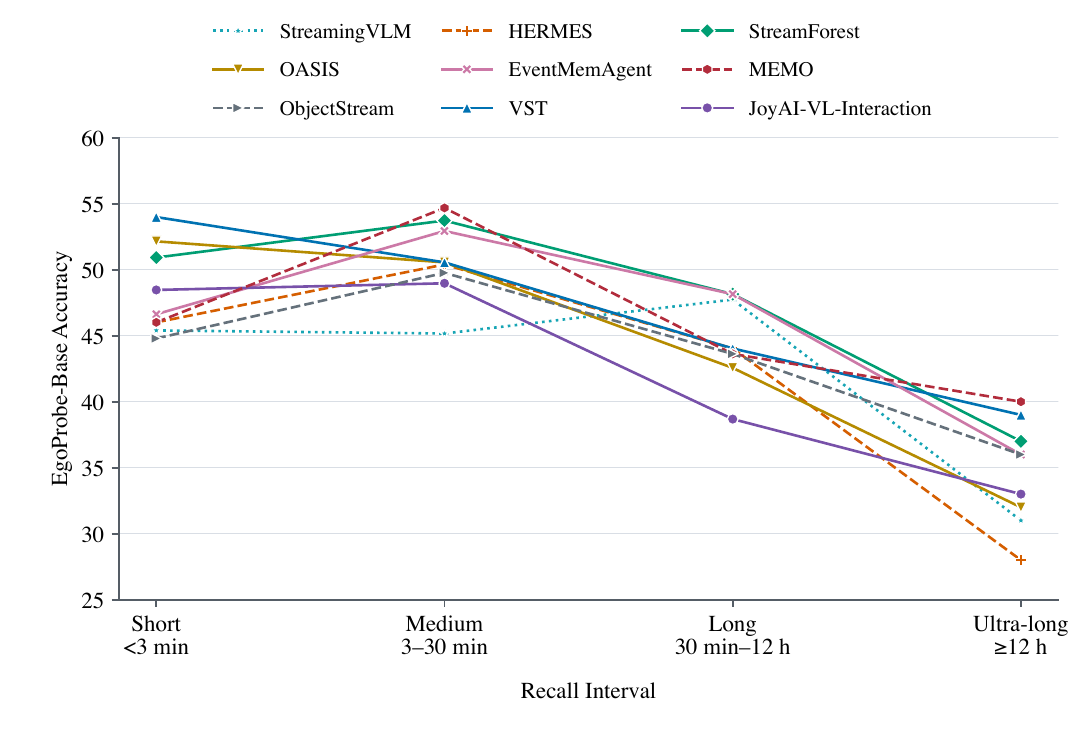}
\setlength{\abovecaptionskip}{2pt}
\caption{EgoProbe-Base accuracy (\%) across recall intervals for the nine models in Table~\ref{tab:temporal_gap}.}
\label{fig:base_recall_interval}
\end{figure}

\section{Additional Experimental Results}
\label{app:results}

\subsection{Recall-Interval Analysis}
\label{app:results_3}
Figure~\ref{fig:base_recall_interval} visualizes the nine model entries in Table~\ref{tab:temporal_gap}. The unweighted mean across these models is 48.26\% in Short and 50.75\% in Medium, falling to 44.49\% in Long and 34.67\% in Ultra-long. All nine models obtain their lowest accuracy in the Ultra-long group. Overall, accuracy tends to decrease as the recall interval grows, showing a pattern resembling a forgetting curve and highlighting the difficulty of retaining information for long-range recall.

\par\vspace{8pt}
\begingroup
\setlength{\intextsep}{4pt}
\setlength{\textfloatsep}{4pt}
\begin{table}[htbp]
\setlength{\abovecaptionskip}{0pt}
\centering\small
\caption{Qwen3.5 single-turn accuracy (\%) with and without video on EgoProbe-Base and EgoProbe-Echo single turn QAs.}
\label{tab:appendix_text_only}
\begin{tabularx}{\linewidth}{@{}Xcc@{}}
\toprule
\textbf{Model / Input} & \textbf{Base Acc.}$\uparrow$ & \textbf{Echo Acc.}$\uparrow$\\
\midrule
Qwen3.5 --- Text-only & 25.51 & 24.49\\
Qwen3.5 --- Video + Text & 45.91 & 35.31\\
\bottomrule
\end{tabularx}
\end{table}

\endgroup

\subsection{Text-Only Reference}
\label{app:results_text}
Table~\ref{tab:appendix_text_only} compares Qwen3.5 with and without visual input on the same EgoProbe-Base and EgoProbe-Echo single-turn questions. Given only the questions and answer options, the model achieves 25.51\% and 24.49\% accuracy, respectively, both close to the 25\% random-choice baseline. This suggests that linguistic plausibility and prior knowledge alone offer limited help in identifying the correct answers for this model. Adding video raises accuracy to 45.91\% on EgoProbe-Base and 35.31\% on EgoProbe-Echo, corresponding to gains of 20.40 and 10.82 percentage points. The improvement on both subsets supports the importance of visual evidence, while the remaining errors show that access to this evidence does not ensure reliable recall. Together, these results support EgoProbe's evaluation of memory for observed experiences, where models must recover information from the video rather than rely on question and option text alone.
\begingroup
\setlength{\intextsep}{4pt}
\setlength{\textfloatsep}{4pt}
\begin{table}[htbp]
\setlength{\abovecaptionskip}{0pt}
\centering\scriptsize
\caption{Fine-grained EgoProbe-Echo single-turn results (\%). Groups count annotated related events supporting different answers. Event err. denotes event-linked error rate. Bold indicates the highest accuracy and lowest error rate among reported models in each group. }
\label{tab:appendix_fine_events}
\setlength{\tabcolsep}{2pt}
\renewcommand{\arraystretch}{1.15}
\begin{tabularx}{\linewidth}{@{}lYY|YY|YY|YY|YY@{}}
\toprule
& \multicolumn{2}{c}{\textbf{1 event}} & \multicolumn{2}{c}{\textbf{2 events}} & \multicolumn{2}{c}{\textbf{3 events}} & \multicolumn{2}{c}{\textbf{4 events}} & \multicolumn{2}{c}{\textbf{$\geq$5 events}}\\
\cmidrule(lr){2-3}\cmidrule(lr){4-5}\cmidrule(lr){6-7}\cmidrule(lr){8-9}\cmidrule(l){10-11}
\multicolumn{1}{c}{\textbf{Model}} & \multicolumn{1}{c}{\textbf{Acc.}$\uparrow$} & \multicolumn{1}{c}{\textbf{Event err.}$\downarrow$} & \multicolumn{1}{c}{\textbf{Acc.}$\uparrow$} & \multicolumn{1}{c}{\textbf{Event err.}$\downarrow$} & \multicolumn{1}{c}{\textbf{Acc.}$\uparrow$} & \multicolumn{1}{c}{\textbf{Event err.}$\downarrow$} & \multicolumn{1}{c}{\textbf{Acc.}$\uparrow$} & \multicolumn{1}{c}{\textbf{Event err.}$\downarrow$} & \multicolumn{1}{c}{\textbf{Acc.}$\uparrow$} & \multicolumn{1}{c}{\textbf{Event err.}$\downarrow$}\\
\midrule
StreamingVLM & 37.08 & 35.47 & 22.41 & 52.28 & 29.60 & 60.00 & 26.87 & \textbf{58.21} & 25.43 & 68.21 \\
HERMES & 36.44 & 36.44 & 31.54 & 46.89 & 37.60 & \textbf{52.80} & 25.37 & 64.18 & 30.64 & 64.74 \\
StreamForest & 43.18 & 38.04 & \textbf{39.83} & \textbf{44.40} & 31.20 & 60.00 & \textbf{28.36} & 59.70 & 31.79 & 63.58 \\
OASIS & 42.54 & 34.19 & 29.88 & 56.02 & \textbf{40.00} & 54.40 & 16.42 & 76.12 & 27.17 & 68.79 \\
EventMemAgent & \textbf{44.46} & \textbf{31.94} & 33.61 & 46.89 & 28.00 & 62.40 & 20.90 & 68.66 & \textbf{32.95} & \textbf{63.01} \\
MEMO & 40.29 & 37.08 & 27.80 & 56.43 & 31.20 & 58.40 & 14.93 & 76.12 & 30.06 & 64.74 \\
ObjectStream & 36.44 & 37.40 & 30.71 & 52.28 & 29.60 & 59.20 & 19.40 & 70.15 & 27.17 & 70.52 \\
VST & 42.38 & 37.72 & 30.29 & 53.11 & 27.20 & 64.80 & 14.93 & 77.61 & 31.79 & 64.16 \\
JoyAI-VL-Interaction & 36.28 & 36.76 & 29.05 & 51.87 & 26.40 & 62.40 & 23.88 & 61.19 & 24.28 & 68.79 \\
\bottomrule
\end{tabularx}
\end{table}

\endgroup

\subsection{Fine-Grained Related-Event Analysis}
\label{app:results_fine}
Table~\ref{tab:appendix_fine_events} expands the event grouping in Table~\ref{tab:occurrence_errors} into 1, 2, 3, 4, and at least 5 related events supporting different answers. The groups contain 623, 241, 125, 67, and 173 questions, respectively. We report single-turn accuracy and event-linked error rate with fixed group denominators.

Every model with reported group-level results has lower accuracy and a higher event-linked error rate in the at-least-five-event group than in the one-event group. For example, StreamForest changes from 43.18\% accuracy and 38.04\% event-linked errors to 31.79\% and 63.58\%, respectively. These results highlight the value of EgoProbe's event-grounded design: the benchmark evaluates target-specific recall across different numbers of related experiences and distinguishes errors linked to those experiences from other incorrect choices. This provides a more detailed view of recall failures than overall accuracy alone.

\end{document}